\documentclass[twoside]{article}

\usepackage[accepted]{aistats2021}
\usepackage[round]{natbib}

\usepackage{times}  
\usepackage{helvet} 
\usepackage{courier} 
\usepackage[hyphens]{url}  
\usepackage{graphicx} 
\usepackage{natbib} 
\usepackage{caption}
\usepackage[utf8]{inputenc}
\usepackage{url}    
\usepackage{booktabs}     
\usepackage{amsfonts}    
\usepackage{nicefrac}   
\usepackage{microtype}  
\usepackage{graphicx}
\usepackage{subcaption}
\usepackage{amsmath,amssymb,amsthm}
\usepackage{algorithm}
\usepackage[noend]{algpseudocode}
\usepackage{listings}
\usepackage{fdsymbol}
\usepackage{mathtools}
\usepackage{dsfont}
\usepackage{enumerate}
\usepackage{makecell}
\usepackage[export]{adjustbox}
\usepackage{tabularx,lipsum,environ,forest,physics,multirow, color, courier, lineno,comment, xspace, colortbl}
\usepackage{soul}
\usepackage{hhline}
\usepackage{letltxmacro}
\usepackage{afterpage}
\makeatletter
\renewcommand{\ALG@beginalgorithmic}{\small}
\makeatother

\newcommand{\nit}[1]{\textcolor{purple}{#1}}

\newcommand{\dc}{DC($\mathcal{B}$)\xspace}

\algdef{SE}[SUBALG]{Indent}{EndIndent}{}{\algorithmicend\ }%
\algtext*{Indent}
\algtext*{EndIndent}

\newtheorem{theorem}{Theorem}
\newtheorem{lemma}[theorem]{Lemma}

\newtheorem{definition}{Definition}
\newtheorem{example}{Example}

\algnewcommand\And{\textbf{and}}
\newcommand{\RN}[1]{%
  \textup{\uppercase\expandafter{\romannumeral#1}}%
}

\usepackage{letltxmacro}
\LetLtxMacro\oldttfamily\ttfamily
\DeclareRobustCommand{\ttfamily}{\oldttfamily\csname ttsize\endcsname}
\newcommand{\setttsize}[1]{\def\ttsize{#1}}%

\setttsize{\small}
\begin{document}

\twocolumn[
\aistatstitle{Context-Specific Likelihood Weighting}
\aistatsauthor{ Nitesh Kumar \And Ond\v rej Ku\v zelka}
\aistatsaddress{ Department of Computer Science and Leuven.AI \\KU Leuven, Belgium \And  Department of Computer Science \\ Czech Technical University in Prague, Czechia}]

\begin{abstract}
Sampling is a popular method for approximate inference when exact inference is impractical. Generally, sampling algorithms do not exploit context-specific independence (CSI) properties of probability distributions. We introduce context-specific likelihood weighting (CS-LW), a new sampling methodology, which besides exploiting the classical conditional independence properties, also exploits CSI properties. Unlike the standard likelihood weighting, CS-LW is based on partial assignments of random variables and requires fewer samples for convergence due to the sampling variance reduction. Furthermore, the speed of generating samples increases. Our novel notion of contextual assignments theoretically justifies CS-LW. We empirically show that CS-LW is competitive with state-of-the-art algorithms for approximate inference in the presence of a significant amount of CSIs.
\end{abstract}

\section{Introduction}
\begin{figure}[t]
    \centering
    \includegraphics[width=1\linewidth]{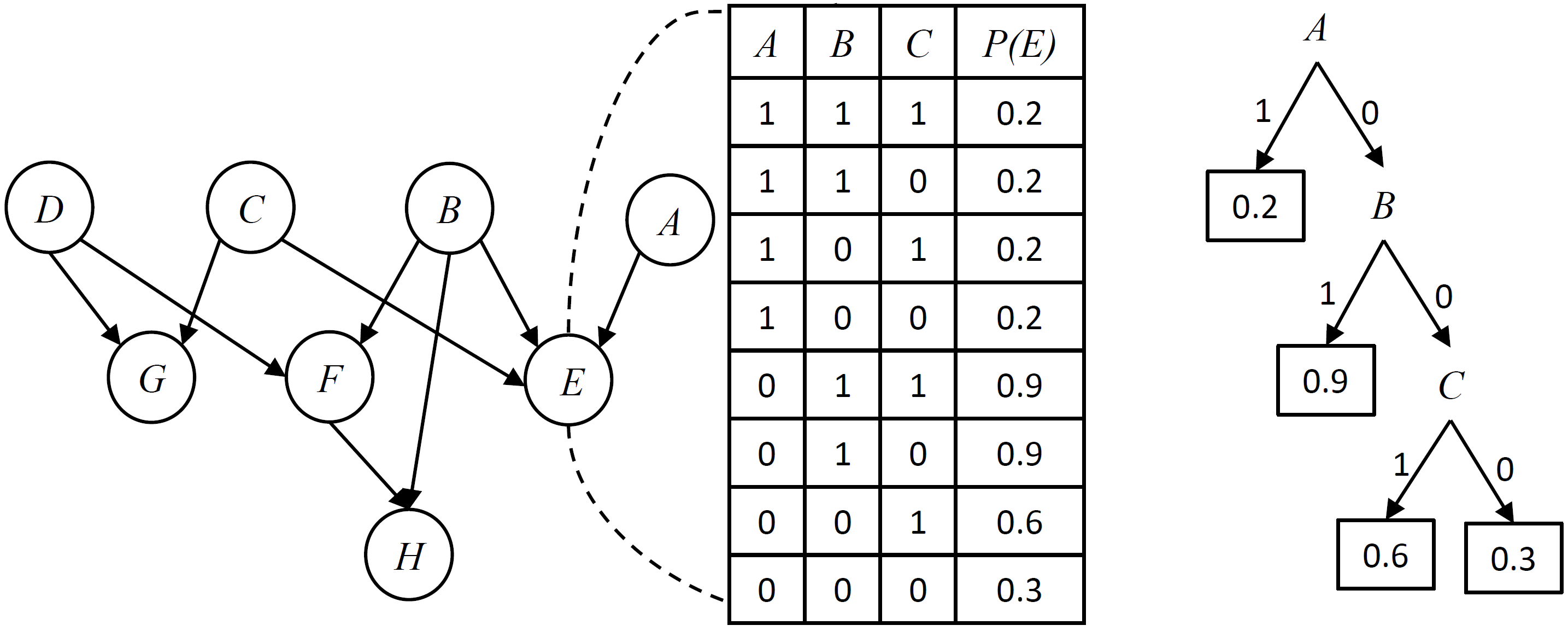}
    \caption{Context-Specific Independence}
    \label{fig:context-specific independence}
\end{figure}

Exploiting independencies present in probability distributions is crucial for feasible probabilistic inference. Bayesian networks (BNs) qualitatively represent {\em conditional independencies} (CIs) over random variables, which allow inference algorithms to exploit them. In many applications, however, exact inference quickly becomes infeasible. The use of stochastic sampling for approximate inference is common in such applications. Sampling algorithms are simple yet powerful tools for inference. They can be applied to arbitrary complex distributions, which is not true for exact inference algorithms. The design of efficient sampling algorithms for BNs has received much attention in the past. Unfortunately, BNs can not represent certain independencies qualitatively: independencies that hold only in certain contexts \citep{boutilier1996context}. These independencies are called {\em context-specific independencies} (CSIs). To illustrate them, consider a BN in Figure \ref{fig:context-specific independence}, where a tree-structure is present in the {\em conditional probability distribution} (CPD) of a random variable $E$. If one observes the CPD carefully, they can conclude that $P(E \mid A=1, B, C) = P(E \mid A=1)$, that is, $P(E \mid A=1, B, C)$ is same for all values of $B$ and $C$. The variable $E$ is said to be independent of variables $\{B,C\}$ in the context $A=1$. These independencies may have global implications, for instance, $E \perp B,C \mid A=1$ implies $E \perp D \mid H, A = 1$. Sampling algorithms generally do not exploit CSIs arising due to structures within CPDs.

One might think that structures in CPDs are accidental. It turns out, however, that such structures are common in many real-world settings. For example, consider a scenario \citep{koller2009probabilistic} where a symptom, \textit{fever}, depends on $10$ diseases. It would be impractical for medical experts to answer $1,024$ questions of the format: ``What is the probability of high fever when the patient has disease $A$ does not have disease $B$ $\dots$?'' It might be the case that, if patients suffer from disease $A$, then they are certain to have a high fever, and our knowledge of their suffering from other diseases does not matter. One might argue, what if we automatically learn BNs from data? In this case, however, a huge amount of data would be needed to learn the parameters that are exponential in the number of parents required to describe the tabular-CPD. The tree-CPDs that require much fewer parameters are a more efficient way of learning BNs automatically from data \citep{chickering1997bayesian,friedman1998bayesian,breese1998empirical}. Moreover, the structures naturally arise due to {\em if-else conditions} in programs written in probabilistic programming languages (PPLs).

There are exact inference algorithms that exploit CSIs, and thus, form state-of-the-art algorithms for exact inference \citep{friedman2018approximate}. These algorithms are based on the knowledge compilation technique \citep{darwiche2003differential} that uses logical reasoning to naturally exploit CSIs. An obvious question, then, is: {\em how to design a sampling algorithm that naturally exploits CSIs, along with CIs?} It is widely believed that CSI properties in distributions are difficult to harness for approximate inference \citep{friedman2018approximate}. In this paper, we 
answer this difficult question by developing a sampling algorithm that can harness both CI and CSI properties. 

To realize this, we adopt {\em likelihood weighting} \citep[LW,][]{shachter1990simulation,fung1990weighing}, a sampling algorithm for BNs; and extend it to a rule-based representation of distributions since rules are known to represent the structures  qualitatively \citep{poole1997probabilistic}.
We call the resulting algorithm {\em context-specific likelihood weighting} (CS-LW) and provide its open-source implementation\footnote{The code is available here: \url{https://github.com/niteshroyal/CS-LW.git}}. Additionally, we present a novel notion of {\em contextual assignments} that provides a theoretical framework for exploiting CSIs.  
Taking advantage of the better representation of structures via rules, CS-LW assigns only a subset of variables required for computing conditional query leading to i) faster convergence, ii) faster speed of generating samples. This contrasts with many modern sampling algorithms such as collapsed sampling, which speed up convergence by sampling only a subset of variables but at the cost of much reduced speed of generating samples. We empirically demonstrate that CS-LW is competitive with state of the art.

\section{Background}
We denote random variables with uppercase letters ($A$) and their assignments with lowercase letters ($a$). Bold letters denote sets ($\mathbf{A}$) and their assignments ($\mathbf{a}$). Parents of the variable $A$ are denoted with $\mathbf{Pa}(A)$ and their assignments with $\mathbf{pa}(A)$. In a probability distribution $P(\mathbf{E},\mathbf{X},\mathbf{Z})$ specified by a Bayesian network $\mathcal{B}$, $\mathbf{E}$ denotes a set of observed variables, $\mathbf{X}$ a set of unobserved query variables and $\mathbf{Z}$ a set of unobserved variable other than query variables. The expected value of $A$ relative to a distribution $Q$ is denoted by $\mathbb{E}_{Q}[A]$. Next, we briefly introduce LW, one of the most popular approximate inference algorithms for BNs. 
\subsection{Likelihood Weighting}\label{section: likelihood weighting}
 A typical query to a probability distribution $P(\mathbf{E},\mathbf{X},\mathbf{Z})$ is to compute $P(\mathbf{x}_q \mid \mathbf{e})$, that is, the probability of $\mathbf{X}$ being assigned $\mathbf{x}_q$ given that $\mathbf{E}$ is assigned $\mathbf{e}$. Following Bayes's rule, we have: 
\begin{equation*}
    \begin{aligned}
    P(\mathbf{x}_q \mid \mathbf{e}) = \frac{P(\mathbf{x}_q, \mathbf{e})}{P(\mathbf{e})} 
    = \frac{\sum_{\mathbf{x}, \mathbf{z}} P(\mathbf{x}, \mathbf{z}, \mathbf{e}) f(\mathbf{x})}{\sum_{\mathbf{x}, \mathbf{z}} P(\mathbf{x}, \mathbf{z}, \mathbf{e})} = \mu, 
    \end{aligned}
\end{equation*}
where $f(\mathbf{x})$ is an indicator function $\mathds{1}\{\mathbf{x} = \mathbf{x}_q\}$, which takes value $1$ when $\mathbf{x} = \mathbf{x}_q$, and $0$ otherwise. We can estimate $\mu$ using LW if we specify $P$ using a Bayesian network $\mathcal{B}$. LW belongs to a family of importance sampling schemes that are based on the observation,
\begin{equation}\label{eq: importance sampling}
    \begin{aligned}
    \mu = \frac{ \sum_{\mathbf{x}, \mathbf{z}} Q(\mathbf{x}, \mathbf{z}, \mathbf{e}) f(\mathbf{x}) (P(\mathbf{x}, \mathbf{z}, \mathbf{e})/Q(\mathbf{x}, \mathbf{z}, \mathbf{e}))}{\sum_{\mathbf{x}, \mathbf{z}} Q(\mathbf{x}, \mathbf{z}, \mathbf{e}) (P(\mathbf{x}, \mathbf{z}, \mathbf{e})/Q(\mathbf{x}, \mathbf{z}, \mathbf{e}))},
    \end{aligned}
\end{equation}
where $Q$ is a {\em proposal distribution} such that $Q>0$ whenever $P>0$. The distribution $Q$ is different from $P$ and is used to draw independent samples. Generally, $Q$ is selected such that the samples can be drawn easily. In the case of LW, to draw a sample, variables $X_i \in \mathbf{X} \cup \mathbf{Z}$ are assigned values drawn from $P(X_i \mid \mathbf{pa}(X_i))$ and variables in $\mathbf{E}$ are assigned their observed values. These variables are assigned in a topological ordering relative to the graph structure of $\mathcal{B}$. Thus, the proposal distribution in the case of LW can be described as follows: 
\begin{equation*}
    \begin{aligned}
    Q(\mathbf{X}, \mathbf{Z}, \mathbf{E}) = \prod_{X_i \in \mathbf{X} \cup \mathbf{Z}} P(X_i \mid \mathbf{Pa}(X_i)) \mid_{\mathbf{E}=\mathbf{e}}
    \end{aligned}.
\end{equation*}
Consequently, it is easy to compute the {\em likelihood ratio}  $P(\mathbf{x}, \mathbf{z}, \mathbf{e})/Q(\mathbf{x}, \mathbf{z}, \mathbf{e})$ in Equation \ref{eq: importance sampling}. All factors in the numerator and denominator of the fraction cancel out except for $P(x_i \mid \mathbf{pa}(X_i))$ where $x_i \in \mathbf{e}$. Thus, 
\begin{equation*}
    \begin{aligned}
    \frac{P(\mathbf{X}, \mathbf{Z}, \mathbf{e})}{Q(\mathbf{X}, \mathbf{Z}, \mathbf{e})} = \prod_{x_i \in \mathbf{e}} P(x_i \mid \mathbf{Pa}(X_i)) = \prod_{x_i \in \mathbf{e}} W_{x_i} = W_{\mathbf{e}}, 
    \end{aligned}
\end{equation*}
where $W_{x_i}$, which is also a random variable, is the {\em weight} of evidence $x_i$. The {\em likelihood ratio} $W_{\mathbf{e}}$ is the product of all of these weights, and thus, it is also a random variable. Given $M$ independent weighted samples from $Q$, we can estimate: 
\begin{equation}\label{equation: Naive Estimate}
    \begin{aligned}
    \hat{\mu} = \frac{\sum_{m=1}^{M} f(\mathbf{x}[m]) w_{\mathbf{e}}[m]}{\sum_{m=1}^{M} w_{\mathbf{e}}[m] }.
    \end{aligned}
\end{equation}

\subsection{Context-Specific Independence}
Next, we formally define the independencies that arise due to the structures within CPDs. 
\begin{definition}
Let $P$ be a probability distribution over variables $\mathbf{U}$, and
let $\mathbf{A}, \mathbf{B}, \mathbf{C}, \mathbf{D}$ be disjoint subsets of $\mathbf{U}$. The variables $\mathbf{A}$ and $\mathbf{B}$ are independent given $\mathbf{D}$ and context $\mathbf{c}$ if $P(\mathbf{A} \mid \mathbf{B}, \mathbf{D}, \mathbf{c}) = P(\mathbf{A} \mid \mathbf{D}, \mathbf{c})$  whenever $P(\mathbf{B}, \mathbf{D}, \mathbf{c}) > 0$. This is denoted by $\mathbf{A} \perp \mathbf{B} \mid \mathbf{D}, \mathbf{c}$. If $\mathbf{D}$ is empty then $\mathbf{A}$ and $\mathbf{B}$ are independent given context $\mathbf{c}$, denoted by $\mathbf{A} \perp \mathbf{B} \mid \mathbf{c}$. 
\end{definition}

Independence statements of the above form are called {\em context-specific independencies} (CSIs). When $\mathbf{A}$ is independent of $\mathbf{B}$ given all possible assignments to $\mathbf{C}$ then we have: $\mathbf{A} \perp \mathbf{B} \mid \mathbf{C}$. The independence statements of this form are generally referred to as {\em conditional independencies} (CIs). Thus, CSI is a more fine-grained notion than CI. The graphical structure in $\mathcal{B}$ can only represent CIs. Any CI can be verified in linear time in the size of the graph. However, verifying any arbitrary CSI has been recently shown to be coNP-hard \citep{corander2019logical}. 


\subsection{Context-Specific CPDs}\label{section: DC into}
A natural representation of the structures in a CPD is via a {\em tree-CPD}, as illustrated in Figure \ref{fig:context-specific independence}. For all assignments to the parents of a variable $ A $, a unique leaf in the tree specifies a (conditional) distribution over $ A $. The path to each leaf dictates the contexts, i.e., {\em partially assigned parents}, given which this distribution is used. It is easier to reason using tree-CPDs if we break them into finer-grained elements. A finer-grained representation of structured CPDs is via rules \citep{poole1997probabilistic, koller2009probabilistic}, where each path from the root to a leaf in each tree-CPD maps to a rule. 
For our purposes, we will use a simple rule-based representation language,  which can be seen as a restricted fragment of {\em Distributional Clauses} \citep[DC,][]{gutmann2011magic}.

\begin{example}\label{example: distributional clauses}
\normalfont A set of rules for the tree-CPD in Figure \ref{fig:context-specific independence}:
\begin{lstlisting}[frame=none]
e ~ bernoulli(0.2) := a=1.
e ~ bernoulli(0.9) := a=0,,b=1.
e ~ bernoulli(0.6) := a=0,,b=0,,c=1.
e ~ bernoulli(0.3) := a=0,,b=0,,c=0.\end{lstlisting}
\end{example}
We can also represent structures in CPDs of discrete-continuous distributions using this form of rules like this:
\begin{example}\label{example: distributional clauses continuous}
\normalfont Consider a machine that breaks down if the cooling of the machine is not working or the ambient temperature is too high. The following set of rules specifies a distribution over \texttt{cool}, \texttt{t}(temperature) and \texttt{broken}, where a CSI is implied: \texttt{broken} is independent of \texttt{cool} in a context \texttt{t>30}.
\begin{lstlisting}[frame=none]
cool ~  bernoulli(0.1).
t ~ gaussian(25,2.2).
broken ~ bernoulli(0.9) := t>30.
broken ~ bernoulli(0.6) := t=<30,,cool=0.
broken ~ bernoulli(0.1) := t=<30,,cool=1.\end{lstlisting}
\end{example}
Intuitively, the {\em head} of a rule (\texttt{h $\sim \mathcal{D} \leftarrow$ b1 $\wedge \dots \wedge$ bn}) defines a random variable \texttt{h}, distributed according to a distribution $\mathcal{D}$, whenever all atoms \texttt{bi} in the {\em body} (an assignment of some parents of the variable) of the rule are true, that is:
$p(\texttt{h} \mid \texttt{b1}, \dots, \texttt{bn}) = \mathcal{D}$.
Since we study tree-CPDs, we focus on {\em mutually exclusive and exhaustive} rules; that is, only one rule for the variable \texttt{h} 
can {\em fire} (each atom in the body of the rule is true) at a time. 
A set of rules forms a {\em program}, which we call the DC($\mathcal{B}$) program.

\begin{definition}
Let $\mathcal{B}$ be a Bayesian network with tree-CPDs specifying a distribution $P$. Let $\mathbb{P}$ be a set of rules such that each path from the root to a leaf of each tree-CPD corresponds to a rule in $\mathbb{P}$. Then $\mathbb{P}$ specifies the same distribution $P$, and $\mathbb{P}$ will be called DC($\mathcal{B}$) program. 
\end{definition}


\section{Exploiting Conditional Independencies}\label{Section: Bayes-ball Simulation}
In this section, we will ignore the structures within CPDs and only exploit the graphical structure of BNs. The approach presented in this section forms the basis of our discussion on CS-LW, where we will also exploit CPDs' structure. 

In Section \ref{section: likelihood weighting}, we used all variables to estimate $\mu$. However, due to CIs, observed states and CPDs of only some variables might be required for computing $\mu$. These variables are called {\em requisite variables}. To get a better estimate of $\mu$, it is recommended to use only these variables. The standard approach is to first apply the Bayes-ball algorithm \citep{shacter1998bayes} over the graph structure in $\mathcal{B}$ to obtain a sub-network of requisite variables, then simulate the sub-network to obtain the weighted samples. An alternative approach that we present next is to use Bayes-ball to simulate the original network $\mathcal{B}$ and focus on only requisite variables to obtain the weighted samples. 



\begin{figure}[t]
    \centering
    \includegraphics[width=1\linewidth]{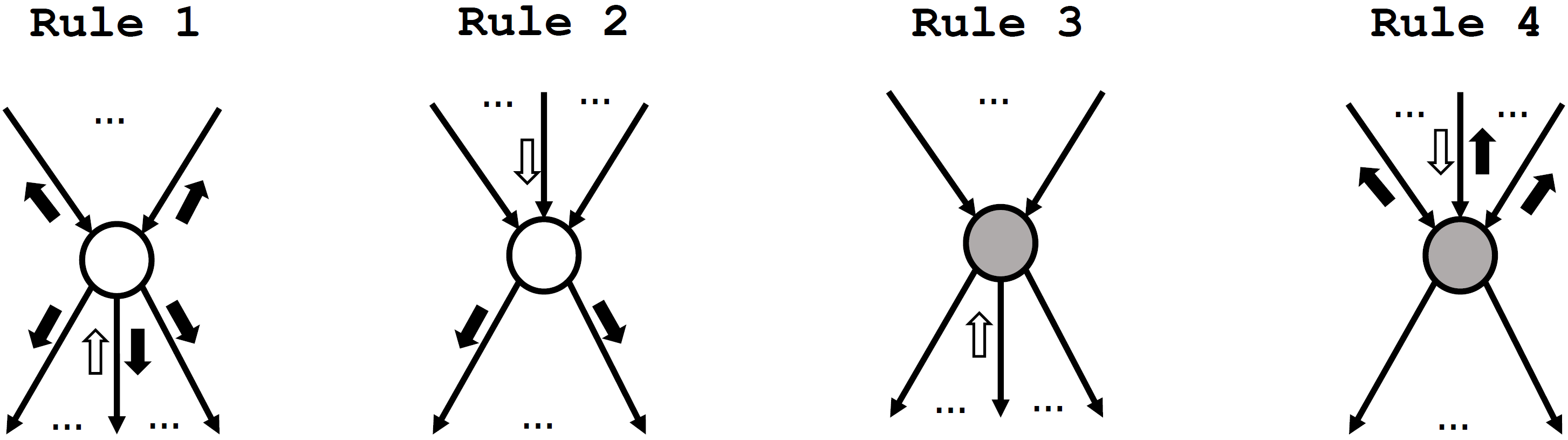}
    \caption{The four rules of Bayes-ball algorithm that decide next visits (indicated using \includegraphics[height=1.75ex]{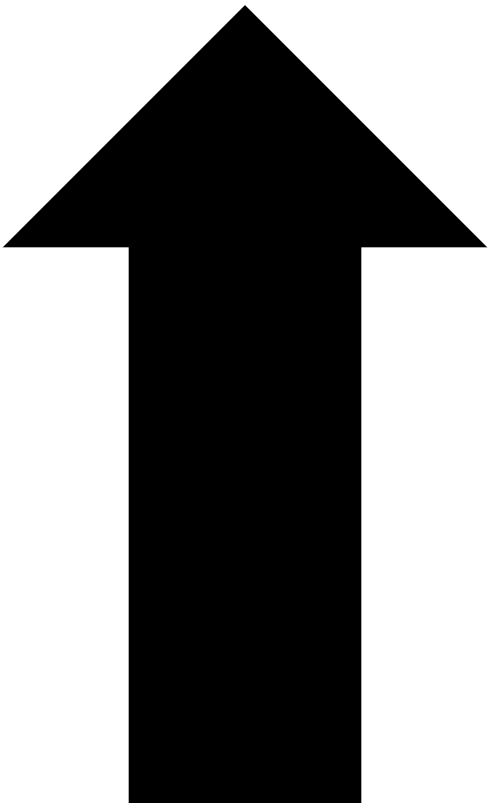}) based on the direction of the current visit (indicated using \includegraphics[height=1.75ex]{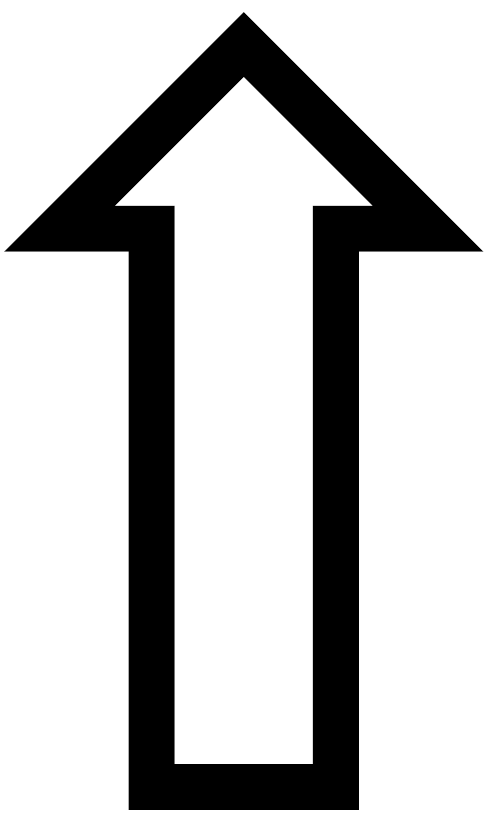}) and the type of variable. To distinguish observed variables from unobserved variables, the former type of variables are shaded.}
    \label{fig: bayes-ball rules}
\end{figure}
To obtain the samples, we need to traverse the graph structure of the Bayesian network $\mathcal{B}$ in a topological ordering. The Bayes-ball algorithm, which is linear in the graph's size, can be used for it. The advantage of using Bayes-ball is that it also detects CIs; thus, it traverses only a sub-graph that depends on the query and evidence. We can also keep assigning unobserved variables, and weighting observed variables along with traversing the graph. In this way, we assign/weigh only requisite variables. The Bayes-ball algorithm uses four rules to traverse the graph (when deterministic variables are absent in $\mathcal{B}$), and marks variables to avoid repeating the same action. These rules are illustrated in Figure \ref{fig: bayes-ball rules}. Next, we discuss these rules and also indicate how to assign/weigh variables, resulting in a new algorithm called {\em Bayes-ball simulation of BNs}. Starting with all query variables scheduled to be visited as if from one of their children, we apply the following rules until no more variables can be visited:
\begin{enumerate}
    \item When the visit of an unobserved variable $U \in \mathbf{X} \cup \mathbf{Z}$ is from a child, and $U$ is not marked on top, then do these in the order: i) Mark $U$ on top; ii) Visit all its parents; iii) Sample a value $y$ from $P(U \mid \mathbf{pa}(U))$ and assign $y$ to $U$; iv) If $U$ is not marked on bottom, then mark $U$ on bottom and visit all its children.  
    \item When the visit of an unobserved variable is from a parent, and the variable is not marked on bottom, then mark the variable on bottom and visit all its children.
    \item When the visit of an observed variable is from a child, then do nothing. 
    \item When the visit of an observed variable $E \in \mathbf{E}$ is from a parent, and $E$ is not marked on top, then do these in the order: i) Mark $E$ on top; ii) Visit all its parents; iii) Let $e$ be a observed value of $E$ and let $w$ be a probability at $e$ according to $P(E \mid \mathbf{pa}(E))$, then the weight of $E$ is $w$.
\end{enumerate}
The above rules define an order for visiting parents and children so that variables are assigned/weighted in a topological ordering. Indeed we can define the order since the original rules for Bayes-ball do not prescribe any order. 
The marks record important information; consequently, we show the following. The proofs for all the results are in the supplementary material. 



\begin{lemma}\label{theorem: bayes-ball 1}
Let $\mathbf{E}_\star \subseteq \mathbf{E}$ be marked on top, $\mathbf{E}_\smwhitestar \subseteq \mathbf{E}$ be visited but not marked on top, and $\mathbf{Z}_\star \subseteq \mathbf{Z}$ be marked on top. Then the query $\mu$ can be computed as follows,
\begin{equation}\label{equation: bb}
    \begin{aligned}
    \mu = 
    \frac{\sum_{\mathbf{x}, \mathbf{z}_\star} P(\mathbf{x},  \mathbf{z}_\star, \mathbf{e}_\star \mid \mathbf{e}_\smwhitestar) f(\mathbf{x})}{\sum_{\mathbf{x}, \mathbf{z}_\star} P(\mathbf{x},  \mathbf{z}_\star, \mathbf{e}_\star \mid \mathbf{e}_\smwhitestar)}
    \end{aligned}
\end{equation}
\end{lemma}
Now, since $\mathbf{X}, \mathbf{Z}_\star, \mathbf{E}_\star, \mathbf{E}_\smwhitestar$ are variables of $\mathcal{B}$ and they form a sub-network $\mathcal{B}_\star$ such that $\mathbf{E}_\smwhitestar$ do not have any parent, we can write,
\begin{equation*}
\begin{aligned}
    P(\mathbf{x}, \mathbf{z}_\star, \mathbf{e}_\star \mid \mathbf{e}_\smwhitestar) = \prod_{u_i \in \mathbf{x} \cup \mathbf{z}_\star \cup \mathbf{e}_\star} P(u_i \mid \mathbf{pa}(U_i))
\end{aligned}
\end{equation*}
such that $\forall p \in \mathbf{pa}(U_i): p \in \mathbf{x} \cup \mathbf{z}_\star \cup \mathbf{e}_\star \cup \mathbf{e}_\smwhitestar$. This means CPDs of some observed variables are not required for computing $\mu$. Now we define these variables.
\begin{definition}
The observed variables whose observed states and CPDs might be required to compute $\mu$ will be called diagnostic evidence. 
\end{definition}

\begin{definition}
The observed variables whose only observed states might be required to compute $\mu$ will be called predictive evidence.  
\end{definition}

Diagnostic evidence (denoted by $\mathbf{e}_\star$) is marked on top, while predictive evidence (denoted by $\mathbf{e}_\smwhitestar$) is visited but not marked on top. The variables $\mathbf{X}$, $\mathbf{Z}_\star$, $\mathbf{E}_\star$, $\mathbf{E}_\smwhitestar$ will be called requisite variables.
Now, we can sample from a factor $Q_\star$ of $Q$ such that,
\begin{equation}\label{equation: proposal distribution}
    \begin{aligned}
    Q_\star(\mathbf{X},  \mathbf{Z}_\star, \mathbf{E}_\star \mid \mathbf{E}_\smwhitestar) = \prod_{X_i \in \mathbf{X} \cup \mathbf{Z}_\star} P(X_i \mid \mathbf{Pa}(X_i)) \mid_{\mathbf{E}_\star=\mathbf{e}_\star}
    \end{aligned}
\end{equation}
When we use Bayes-ball, precisely this factor is considered for sampling. Starting by first setting $\mathbf{E}_\smwhitestar$ their observed values, $\mathbf{X} \cup \mathbf{Z}_\star$ is assigned and $\mathbf{e}_\star$ is weighted in the topological ordering. Given $M$ weighted samples $\mathcal{D}_{\star} = \langle \mathbf{x}[1], w_{\mathbf{e}_{\star}}[1]\rangle, \dots, \langle\mathbf{x}[M],  w_{\mathbf{e}_{\star}}[M] \rangle$ from $Q_\star$, we can estimate:
\begin{equation}\label{Equation: Bayes-ball Estimate}
    \begin{aligned}
    \tilde{\mu} = \frac{ \sum_{m=1}^{M} f(\mathbf{x}[m]) w_{\mathbf{e_\star}}[m]}{\sum_{m=1}^{M} w_{\mathbf{e_\star}}[m]}.
    \end{aligned}
\end{equation}

In this way, we sample from a lower-dimensional space; thus, the new estimator $\tilde{\mu}$ has a lower variance compared to $\hat{\mu}$ due to the Rao-Blackwell theorem.
Consequently, fewer samples are needed to achieve the same accuracy. Hence, we exploit CIs using the graphical structure in $\mathcal{B}$ for improved inference.

\section{Exploiting CSIs}
Now, we will exploit the graphical structure as well as structures within CPDs. This section is divided into two parts. The first part presents a novel notion of contextual assignments that forms a theoretical framework for exploiting CSIs. It provides an insight into the computation of $\mu$ using partial assignments of requisite variables. We will show that CSIs allow for breaking the main problem of computing $\mu$ into several sub-problems that can be solved independently. The second part presents CS-LW based on the notion introduced in the first part, where we will exploit the structure of rules in the program to sample variables given the states of only some of their requisite ancestors. This contrasts with our discussion till now for BNs where knowledge of all such ancestors' state is required.


\subsection{Notion of Contextual Assignments}

We will consider the variables $\mathbf{X}$, $\mathbf{Z}_\star$, $\mathbf{E}_\star$, $\mathbf{E}_\smwhitestar$ requisite for computing the query $\mu$ to the distribution $P$ and the sub-network $\mathcal{B}_\star$ formed by these variables. We start by 
defining the partial assignments that we will use to compute $\mu$ at the end of this section.
\begin{definition} 
Let $\mathbf{Z}_\dagger \subseteq \mathbf{Z}_\star$ and $\mathbf{e}_\dagger \subseteq \mathbf{e}_\star$. Denote $\mathbf{Z}_\star \setminus \mathbf{Z}_\dagger$ by $\mathbf{Z}_\ddagger$, and $\mathbf{e}_\star \setminus \mathbf{e}_\dagger$ by $\mathbf{e}_\ddagger$.
A partial assignment $\mathbf{x}$, $\mathbf{z}_\dagger$, $\mathbf{Z}_\ddagger$,  $\mathbf{e}_\dagger$, $\mathbf{e}_\ddagger$ will be called contextual assignment if due to CSIs in $P$,
\begin{equation*}
    \prod_{u_i \in \mathbf{x} \cup \mathbf{z}_\dagger \cup \mathbf{e}_\dagger} P(u_i \mid \mathbf{pa}(U_i)) = \prod_{u_i \in \mathbf{x} \cup \mathbf{z}_\dagger \cup \mathbf{e}_\dagger} P(u_i \mid \mathbf{ppa}(U_i))
\end{equation*}
where $\mathbf{ppa}(U_i) \subseteq \mathbf{pa}(U_i)$ is a set of partially assigned parents of $U_i$ such that $\mathbf{Z}_\ddagger \cap \mathbf{Ppa}(U_i) = \emptyset$.
\end{definition}

\begin{example}\label{example: contexual assignment}
Consider the network of Figure \ref{fig:context-specific independence}, and assume that our diagnostic evidence is $\{F=1, G=0, H=1\}$, predictive evidence is $\{D=1\}$, and query is $\{E=0\}$. From the CPD's structure, we have: $P(E=0 \mid A=1, B, C) = P(E=0 \mid A=1)$; consequently, a contextual assignment is $\mathbf{x} = \{E=0\}, \mathbf{z}_\dagger = \{A=1\}, \mathbf{e}_\dagger= \{\}, \mathbf{Z}_\ddagger = \{B,C\}, \mathbf{e}_\ddagger=\{F=1,G=0,H=1\}$. We also have: $P(E=0 \mid A=0, B=1, C) = P(E=0 \mid A=0, B=1)$; consequently, another such assignment is $\mathbf{x} = \{E=0\}, \mathbf{z}_\dagger = \{A=0,B=1\}, \mathbf{e}_\dagger= \{H=1\}, \mathbf{Z}_\ddagger = \{C\}, \mathbf{e}_\ddagger=\{F=1,G=0\}$.
\end{example}
We aim to treat the evidence $\mathbf{e}_\ddagger$ independently, thus, we define it first.
\begin{definition}\label{Definition: residual diagnostic evidence} 
The diagnostic evidence $\mathbf{e}_\ddagger$ in a contextual assignment  $\mathbf{x}$, $\mathbf{z}_\dagger$, $\mathbf{Z}_\ddagger$,  $\mathbf{e}_\dagger$, $\mathbf{e}_\ddagger$ will be called residual evidence.
\end{definition}

\begin{figure}[t]
    \centering
    \includegraphics[width=1\linewidth]{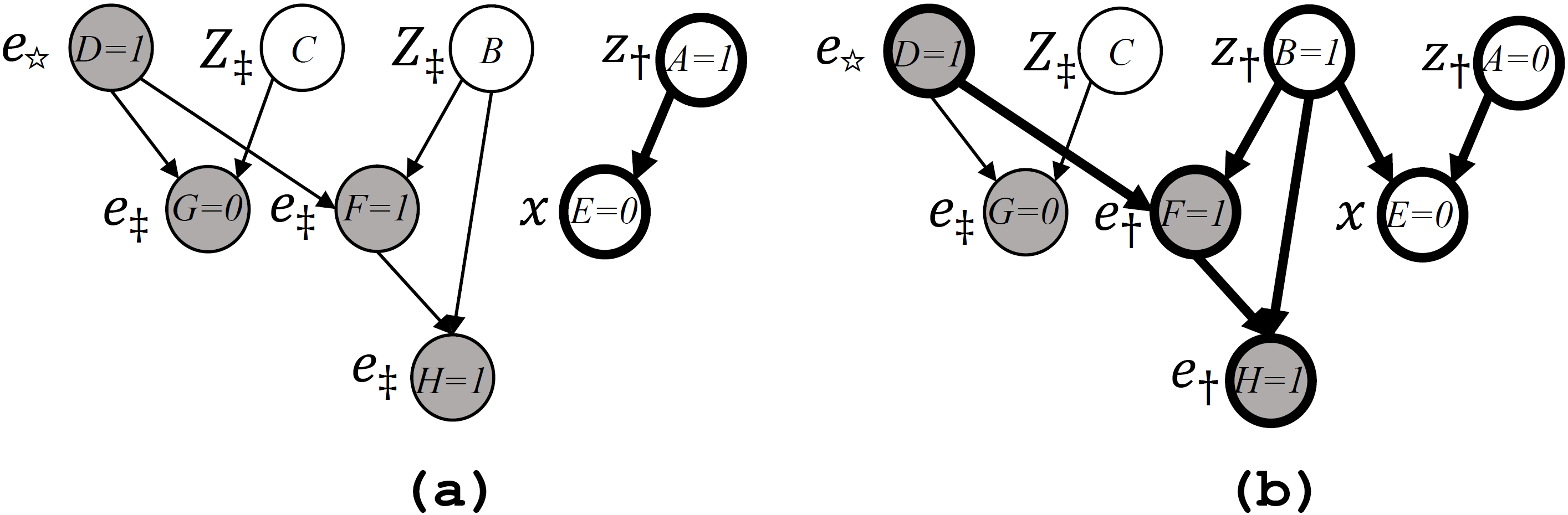}
    \caption{Two safe contextual assignments to variables of BN in Figure \ref{fig:context-specific independence}: (a) in the context ${A=1}$, where edges $C \rightarrow E$ and $B \rightarrow E$ are redundant since $E \perp B, C \mid A=1$; (b) in the context ${A=0, B=1}$, where the edge $C \rightarrow E$ is redundant since $E \perp C \mid A=0, B=1$. To identify such assignments, intuitively, we should apply the Bayes-ball algorithm after removing these edges. Portions of graphs that the algorithm visits, starting with visiting the variable $E$ from its child, are highlighted. Notice that variables $\mathbf{x}, \mathbf{z}_\dagger, \mathbf{e}_\dagger$ lie in the highlighted portion.}
    \label{fig: sub-graphs}
\end{figure}

However, contextual assignments do not immediately allow us to treat the residual evidence independently.  We need the assignments to be safe. 
\begin{definition}\label{definition: Basis} 
Let $e \in \mathbf{e}_\star$ be a diagnostic evidence, and let $S$ be an unobserved ancestor of $E$ in the graph structure in $\mathcal{B}_\star$, where $\mathcal{B}_\star$ is the sub-network formed by the requisite variables. Let $S \rightarrow \cdots\ B_i\ \cdots \rightarrow E$ be a causal trail such that either no $B_i$ is observed or there is no $B_i$. Let $\mathbf{S}$ be a set of all such $S$. Then the variables $\mathbf{S}$ will be called basis of $e$. Let $\mathbf{\dot{e}}_\star \subseteq \mathbf{e}_\star$, and let $\mathbf{\dot{S}}_{\star}$ be a set of all such $S$ for all $e \in \mathbf{\dot{e}}_\star$. Then $\mathbf{\dot{S}}_{\star}$ will be called basis of $\mathbf{\dot{e}}_\star$.
\end{definition}
Reconsider Example \ref{example: contexual assignment}; the basis of $\{F=1\}$ is $\{B\}$. 
\begin{definition}\label{Definition: safe contexual assignments} 
Let $\mathbf{x}$, $\mathbf{z}_\dagger$, $\mathbf{Z}_\ddagger$,  $\mathbf{e}_\dagger$, $\mathbf{e}_\ddagger$ be a contextual assignment, and let $\mathbf{S}_{\ddagger}$ be the basis of the residual evidence $\mathbf{e}_\ddagger$. If $\mathbf{S}_\ddagger \subseteq \mathbf{Z}_\ddagger$ then the contextual assignment will be called safe.
\end{definition}
\begin{example}
Reconsider Example \ref{example: contexual assignment}; the first example of contextual assignment is safe, but the second is not since the basis $B$ of $\mathbf{e}_\ddagger$ is assigned in $\mathbf{z}_\dagger$. We can make the second safe like this: $\mathbf{x} = \{E=0\}, \mathbf{z}_\dagger = \{A=0,B=1\}, \mathbf{e}_\dagger= \{F=1,H=1\}, \mathbf{Z}_\ddagger = \{C\}, \mathbf{e}_\ddagger=\{G=0\}$. See Figure \ref{fig: sub-graphs}.
\end{example}
Before showing that the residual evidence can now be treated independently, we first define a random variable called {\em weight}. 
\begin{definition} 
Let $e \in \mathbf{e}_\star$ be a diagnostic evidence, and let $W_e$ be a random variable defined as follows:
\begin{equation*}
    W_e = P(e \mid \mathbf{Pa}(E)).
\end{equation*}
The variable $W_e$ will be called weight of $e$. The weight of a subset $\mathbf{\dot{e}}_\star \subseteq \mathbf{e}_\star$ is defined as follows:
\begin{equation*}
    W_{\mathbf{\dot{e}}_\star} = \prod_{u_i \in \mathbf{\dot{e}}_\star} P(u_i \mid \mathbf{Pa}(U_i)).
\end{equation*}
\end{definition}

Now we can show the following result:
\begin{theorem}\label{Theorem: the expecation}
Let $\mathbf{\dot{e}}_{\star} \subseteq \mathbf{e}_\star$, and let $\mathbf{\dot{S}}_{{\star}}$ be the basis of $\mathbf{\dot{e}}_{\star}$. Then the expectation of weight $W_{\mathbf{\dot{e}}_{\star}}$ relative to the distribution $Q_\star$ as defined in Equation \ref{equation: proposal distribution} can be written as:
\begin{equation*}
\begin{aligned}
    \mathbb{E}_{Q_\star}[W_{\mathbf{\dot{e}}_{\star}}] = \sum_{\mathbf{\dot{s}}_{\star}} \prod_{u_i \in \mathbf{\dot{e}}_{\star} \cup \mathbf{\dot{s}}_{\star}} P(u_i \mid \mathbf{pa}(U_i)).
\end{aligned}
\end{equation*}
\end{theorem}
Hence, apart from unobserved variables $\mathbf{\dot{S}}_{{\star}}$, the computation of $\mathbb{E}_{Q_\star}[W_{\mathbf{\dot{e}}_{\star}}]$ does not depend on other unobserved variables. Now we are ready to show our main result:


\begin{theorem}\label{theorem: cslw_main}
Let $\Psi$ be a set of all possible safe contextual assignments in the distribution $P$. Then the query $\mu$ to $P$ can be computed as follows:
\begin{equation}
    \begin{aligned}
    \frac{\sum\limits_{\psi \in \Psi} \bigl(\prod\limits_{u_i \in \mathbf{x}[\psi] \cup \mathbf{z}_\dagger[\psi] \cup \mathbf{e}_\dagger[\psi]} P(u_i \mid \mathbf{ppa}(U_i)) f(\mathbf{x}[\psi]) R[\psi]\bigr)}{\sum\limits_{\psi \in \Psi} \bigl(\prod\limits_{u_i \in \mathbf{x}[\psi] \cup \mathbf{z}_\dagger[\psi] \cup \mathbf{e}_\dagger[\psi]} P(u_i \mid \mathbf{ppa}(U_i)) R[\psi])\bigr)}
    \end{aligned}
\end{equation}
where $R[\psi]$ denotes $\mathbb{E}_{Q_\star}[W_{\mathbf{e}_\ddagger[\psi]}]$.
\end{theorem}

We draw some important conclusions: i) $\mu$ can be exactly computed by performing the summation over all safe contextual assignments; notably, variables in $\mathbf{Z}_\dagger$ vary, and so does variables in $\mathbf{E}_\dagger$; ii) For all $\psi \in \Psi$, the computation of $\mathbb{E}_{Q_\star}[W_{\mathbf{e}_\ddagger[\psi]}]$ does not depend on the context $\mathbf{x[\psi]}, \mathbf{z}_\dagger[\psi]$ since no basis of $\mathbf{e}_\ddagger[\psi]$ is assigned in the context (by Theorem \ref{Theorem: the expecation}).
Hence, $\mathbb{E}_{Q_\star}[W_{\mathbf{e}_\ddagger[\psi]}]$ can be computed independently. However, the context 
decides which evidence should be in the subset $\mathbf{e}_\ddagger[\psi]$. That is why we can not cancel $\mathbb{E}_{Q_\star}[W_{\mathbf{e}_\ddagger[\psi]}]$ from the numerator and denominator. 

\subsection{Context-Specific Likelihood Weighting}
First, we present an algorithm that simulates the \dc program $\mathbb{P}$, specifying the same distribution $P$, to generate safe contextual assignments. Then we discuss how to estimate the expectations independently before estimating $\mu$.
\subsubsection{Simulation of \dc Programs}
We start by asking a question. Suppose we modify the first and the fourth rule of Bayes-ball simulation, introduced in Section \ref{Section: Bayes-ball Simulation}, as follows:
\begin{itemize}
    \item In the first rule, when the visit of an unobserved variable is from its child, everything remains the same except that only \underline{some parents} are visited, not all.
    \item Similarly, in the fourth rule, when the visit of an observed variable is from its parent, everything remains the same except that only \underline{some parents} are visited.
\end{itemize}
\begin{algorithm}[t]
\caption{Simulation of \dc Programs}
\label{algorithm: forward-backward}
\begin{algorithmic}
\Procedure{Simulate-DC}{$\mathbb{P}, \mathbf{x}, \mathbf{e}$}
\begin{itemize}
    \item Visits variables from parent and also simulates a \dc program $\mathbb{P}$ based on inputs: i) $\mathbf{x}$, a query; ii) $\mathbf{e}$: evidence. 
    \item Output: i) \texttt{i}: $f(\mathbf{x})$ that can be either $0$ or $1$; ii) \texttt{W}: a table of weights of diagnostic evidence ($\mathbf{e}_\dagger$). 
    \item The procedure maintains global data structures: i) \texttt{Asg}, a table that records assignments of variables ($\mathbf{x} \cup \mathbf{z}_\dagger$); ii) \texttt{Dst}, a table that records distributions for variables; iii) \texttt{Forward}, a set of variables whose children to be visited from parent; iv) \texttt{Top}, a set of variables marked on top; v) \texttt{Bottom}, a set of variables marked on bottom.
\end{itemize}
\begin{enumerate}
    \item Empty \texttt{Asg}, \texttt{Dst}, \texttt{W}, \texttt{Top}, \texttt{Bottom}, \texttt{Forward}.
    \item If \Call{prove-marked}{$\mathbf{x}$}\texttt{==yes} then  $\texttt{i}=1$ else $\texttt{i}=0$.
    \item While \texttt{Forward} is not empty:
    \begin{enumerate}
        \item Remove \texttt{m} from \texttt{Forward}.
        \item For all \texttt{h $\sim \mathcal{D} \leftarrow $ Body} in $\mathbb{P}$ such that \texttt{m=z} in \texttt{Body}:
        \begin{enumerate}
            \item If \texttt{h} is observed in $\mathbf{e}$ and \texttt{h} not in \texttt{Top}:
            \begin{enumerate}
                \item Add \texttt{h} to \texttt{Top}
                \item For all \texttt{h $\sim \mathcal{D} \leftarrow$ Body} in $\mathbb{P}$: \Call{Prove-marked}{\texttt{Body} $\wedge$ \texttt{dist(h,$\mathcal{D}$)}}.
                \item Let \texttt{x} be a observed value of \texttt{h} and let \texttt{p} be a probability at \texttt{x} according to distribution \texttt{Dst[h]}. Record \texttt{W[h]=p}.
            \end{enumerate}
            \item If \texttt{h} is not observed in $\mathbf{e}$ and \texttt{h} not in \texttt{Bottom}: 
            \begin{enumerate}
                \item Add \texttt{h} to \texttt{Bottom} and add \texttt{h} to \texttt{Forward}.
            \end{enumerate}
        \end{enumerate}
    \end{enumerate}
    \item Return \texttt{[i,W]}.
\end{enumerate}
\EndProcedure
\end{algorithmic}
\end{algorithm}
Which variables will be assigned, and which will be weighted using the modified simulation rules? Intuitively, only a subset of variables in $\mathbf{Z}_\star$ should be assigned, and only a subset of variables in $\mathbf{E}_\star$ should be weighted. But then how to assign/weigh a variable knowing the state of only some of its parent. We can do
that when structures are present in CPDs, and these structures are explicitly represented using rules, as discussed in Section \ref{section: DC into}. This is because rules define the distribution from which the variable should be sampled, although the state of some parents of the variable is known before that. Hence, the key idea is to visit only some parents (if possible due to structures); consequently, those unobserved parents that are not visited might not be required to be sampled. 

To realize that, we need to modify the Bayes-ball simulation such that it works on \dc programs. This modified simulation for \dc programs is defined procedurally in Algorithm \ref{algorithm: forward-backward}. The algorithm visits variables from their parents and calls Algorithm \ref{algorithm: proof} to visit variables from their children. Like Bayes-ball, this algorithm also marks variables on top and bottom to avoid repeating the same action. Readers familiar with theorem proving will find that Algorithm \ref{algorithm: proof} closely resembles SLD resolution \citep{kowalski1974predicate}, but it is also different since it is stochastic. An example illustrating how Algorithm \ref{algorithm: proof} visits only some requisite ancestors to sample a variable is present in the supplementary material. 

\begin{algorithm}[t]
\caption{\dc Proof Procedure}
\label{algorithm: proof}
\begin{algorithmic}
\Procedure{prove-marked}{\texttt{Goal}}
\begin{itemize}
    \item Visits variables from child, consequently, proves a conjunction of atoms \texttt{Goal}. Returns \texttt{yes}; otherwise fails.
    \item Accesses the program $\mathbb{P}$, the set \texttt{Top}, the tables \texttt{Asg}, \texttt{Dst} and evidence $\mathbf{e}$ as defined in Algorithm \ref{algorithm: forward-backward}. 
\end{itemize}
\begin{enumerate}
    \item While \texttt{Goal} in not empty:
    \begin{enumerate}
        \item Select the first atom \texttt{b} from \texttt{Goal}.
        \item If \texttt{b} is of the form \texttt{a=x}:
        \begin{enumerate}
            \item If \texttt{a} is observed in $\mathbf{e}$ then let \texttt{y} is the value of \texttt{a}.
            \item Else if \texttt{a} in \texttt{Top} then \texttt{y=Asg[a]}.
            \item Else:
            \begin{enumerate}
                \item Add \texttt{a} to \texttt{Top}.
                \item For all \texttt{a $\sim \mathcal{D} \leftarrow$ Body} in $\mathbb{P}$: \Call{Prove-marked}{\texttt{Body} $\wedge$ \texttt{dist(a,$\mathcal{D}$)}}
                \item Sample a value \texttt{y} from distribution \texttt{Dst[a]} and record \texttt{Asg[a]=y}. 
                \item If \texttt{a} not in \texttt{Bottom}: add \texttt{a} to \texttt{Bottom} and add \texttt{a} to \texttt{Forward}.
            \end{enumerate}
            \item If \texttt{x==y} then remove \texttt{b} from \texttt{Goal} else fail.
        \end{enumerate}
            \item If \texttt{b} is of the form \texttt{dist(a,$\mathcal{D}$)}: record \texttt{Dst[a]=$\mathcal{D}$} and remove \texttt{b} from \texttt{Goal}.
    \end{enumerate}
    \item Return \texttt{yes}.
\end{enumerate}
\EndProcedure
\end{algorithmic}
\end{algorithm}

Since the simulation of $\mathbb{P}$ follows the same four rules of Bayes-ball simulation except that only some parents are visited in the first and fourth rule, we show that 
\begin{lemma}\label{theorem: partial assignments}
Let $\mathbf{E_{\dagger}}$ be a set of observed variables weighed and let $\mathbf{Z_{\dagger}}$ be a set of unobserved variables, apart from query variables, assigned in a simulation of $\mathbb{P}$, then,
\begin{equation*}
    \mathbf{Z_{\dagger}} \subseteq \mathbf{Z_{\star}} \text{ and } \mathbf{E_{\dagger}} \subseteq \mathbf{E_{\star}}.
\end{equation*}
\end{lemma}
The query variables $\mathbf{X}$ are always assigned since the simulation starts with visiting these variables as if visits are from one of their children. To simplify notation, from now on we use $\mathbf{Z}_\dagger$ to denote the subset of variables in $\mathbf{Z}_\star$ that are assigned, $\mathbf{E}_\dagger$ to denote the subset of variables in $\mathbf{E}_\star$ that are weighted in the simulation of $\mathbb{P}$. $\mathbf{Z}_\ddagger$ to denote $\mathbf{Z}_\star \setminus \mathbf{Z}_\dagger$,
and $\mathbf{E}_\ddagger$ to denote $\mathbf{E}_\star \setminus \mathbf{E}_\dagger$
We show that the simulation performs safe contextual assignments to requisite variables. 
\begin{theorem}\label{theorem: dc-partial-justification}
The partial assignment $\mathbf{x}$, $\mathbf{z}_\dagger$, $\mathbf{Z}_\ddagger$,  $\mathbf{e}_\dagger$, $\mathbf{e}_\ddagger$ generated in a simulation of $\mathbb{P}$ is a safe contextual assignment. 
\end{theorem}
The proof of Theorem \ref{theorem: dc-partial-justification} relies on the following Lemma. 

\begin{lemma}\label{theorem: DC CSI}
Let $\mathbb{P}$ be a \dc program specifying a distribution $P$. Let $\mathbf{B}, \mathbf{C}$ be disjoint sets of parents of a variable $A$. In the simulation of $\mathbb{P}$, if $A$ is sampled/weighted, given an assignment $\mathbf{c}$, and without assigning $\mathbf{B}$, then,
\begin{equation*}
P(A \mid \mathbf{c}, \mathbf{B}) = P(A \mid \mathbf{c}).
\end{equation*}
\end{lemma}


Hence, just like the standard LW, we sample from a factor $Q_\dagger$ of the proposal distribution $Q_\star$, which is given by, 
\begin{equation*}
    \begin{aligned}
    Q_\dagger = \prod_{u_i \in \mathbf{x} \cup \mathbf{z}_\dagger \cup \mathbf{e}_\dagger} P(u_i \mid \mathbf{ppa}(U_i))
    \end{aligned}
\end{equation*}
where $P(u_i \mid \mathbf{ppa}(U_i)) = 1$ if $u_i \in \mathbf{e}_\dagger$.
It is precisely this factor that Algorithm \ref{algorithm: forward-backward} considers for the simulation of $\mathbb{P}$. Starting by first setting $\mathbf{E}_\smwhitestar$, $\mathbf{E}_\ddagger$ their observed values, it assigns $\mathbf{X} \cup \mathbf{Z}_\dagger$ and weighs $\mathbf{e}_\dagger$ in the topological ordering. In this process, 
it records {\em partial weights} $\mathbf{w}_{\mathbf{e}_\dagger}$, such that: $\prod_{x_i \in \mathbf{e}_\dagger} w_{x_i} = w_{\mathbf{e}_\dagger}$ and $w_{x_i} \in \mathbf{w}_{\mathbf{e}_\dagger}$. Given $M$ partially weighted samples $\mathcal{D}_{\dagger} = \langle \mathbf{x}[1], \mathbf{w}_{\mathbf{e}_{\dagger}[1]}\rangle, \dots, \langle\mathbf{x}[M],  \mathbf{w}_{\mathbf{e}_{\dagger}[M]} \rangle$ from $Q_\dagger$, we could estimate $\mu$ using Theorem \ref{theorem: cslw_main} as follows:
\begin{equation}\label{equation: forward-backward sampling}
    \begin{aligned}
    \overline{\mu} = \frac{ \sum_{m=1}^{M} f(\mathbf{x}[m])\times w_{\mathbf{e_\dagger}[m]}\times\mathbb{E}_{Q_\star}[W_{\mathbf{e}_\ddagger[m]}] }{\sum_{m=1}^{M} w_{\mathbf{e_\dagger}[m]}\times \mathbb{E}_{Q_\star}[W_{\mathbf{e}_\ddagger[m]}]}
    \end{aligned}
\end{equation}

However, we still can not estimate it since we still do not have expectations $\mathbb{E}_{Q_\star}[W_{\mathbf{e}_\ddagger[m]}]$. 
Fortunately, there are ways to estimate them from partial weights in $\mathcal{D}_\dagger$. We discuss one such way next. 

\subsubsection{Estimating the Expected Weight of Residuals}
We start with the notion of sampling mean. Let $\mathcal{W}_\star = \langle w_{e_1}[1], \dots, w_{e_m}[1]\rangle, \dots, \langle w_{e_1}[n], \dots, w_{e_m}[n]\rangle$ be a data set of $n$ observations of weights of $m$ diagnostic evidence drawn using the standard LW. How can we estimate the expectation $\mathbb{E}_{Q_\star}[W_{e_i}]$ from $\mathcal{W}_\star$? The standard approach is to use the sampling mean: $\overline{W}_{e_i} = \frac{1}{n}\sum_{r=1}^{n}w_{e_i}[r]$. In general, $\mathbb{E}_{Q_\star}[W_{e_i}\dots W_{e_j}]$ can be estimated using the estimator: $\overline{W_{e_i}\dots W}_{e_j} = \frac{1}{n}\sum_{r=1}^{n}w_{e_i}[r]\dots w_{e_j}[r]$. Since LW draws are independent and identical distributed (i.i.d.), it is easy to show that the estimator is unbiased. 

However, some entries, i.e., weights of residual evidence, are missing in the data set $\mathcal{W}_\dagger$ obtained using CS-LW. The trick is to fill the missing entries by drawing samples of the missing weights once we obtain $\mathcal{W}_\dagger$. More precisely, missing weights $\langle W_{e_i}, \dots, W_{e_j}\rangle$ in $r^{\text{th}}$ row of $\mathcal{W}_\dagger$ are filled in with a joint state $\langle w_{e_i}[r], \dots, w_{e_j}[r] \rangle$ of the weights. To draw the joint state, we again use Algorithm \ref{algorithm: forward-backward} and visit observed variables $\langle E_i, \dots, E_j\rangle$ from parent. 
Once all missing entries are filled in, we can estimate $\mathbb{E}_{Q_\star}[W_{e_i}\dots W_{e_j}]$ using the estimator $\overline{W_{e_i}\dots W}_{e_j}$ as just discussed. Once we estimate all required expectations, it is straightforward to estimate $\mu$ using Equation \ref{equation: forward-backward sampling}.

At this point, we can gain some insight into the role of CSIs in sampling. They allow us 
to estimate the expectation $\mathbb{E}_{Q_\star}[W_{\mathbf{e}_\ddagger}]$ separately. We estimate it from all samples obtained at the end of the sampling process, thereby reducing the contribution $W_{\mathbf{e}_\ddagger}$ makes to the variance of our main estimator $\overline{\mu}$. The residual evidence $\mathbf{e}_\ddagger$ would be large if much CSIs are present in the distribution; consequently, we would obtain a much better estimate of $\mu$ using significantly fewer samples. Moreover, drawing a single sample would be faster since only a subset of requisite variables is visited. Hence, in addition to CIs, we exploit CSIs and improve LW further. We observe all these speculated improvements in our experiments. 

\section{Empirical Evaluation}
We answer three questions empirically:

\textbf{Q1}: How does the sampling speed of CS-LW compare with the standard LW in the presence of CSIs?

\textbf{Q2}: How does the accuracy of the estimate obtained using CS-LW compare with the standard LW ? 

\textbf{Q3}: How does CS-LW compare to the state-of-the-art approximate inference algorithms?  

To answer the first two questions, we need BNs with structures present within CPDs. Such BNs, however, are not readily available since the structure while designing inference algorithms is generally overlooked. We identified two BNs from the Bayesian network repository \citep{bnrepository}, which have many structures within CPDs: i) \textit{Alarm}, a monitoring system for patients with 37 variables; ii) \textit{Andes}, an intelligent tutoring system with 223 variables.

\begin{table}[t]
\centering
\begin{adjustbox}{width=\columnwidth,center}
\begin{tabular}{|c|c|cc|cc|}
\toprule
         &   &   \multicolumn{2}{|c|}{\textbf{LW}} &   \multicolumn{2}{|c|}{\textbf{CS-LW}} \\
\midrule
 \textbf{BN} & \textbf{N} & \textbf{MAE $\pm$ Std.} & \textbf{Time} & \textbf{MAE $\pm$ Std.} & \textbf{Time} \\
\midrule
\midrule
\multirow{4}{*}{\textit{Alarm}} & 100 & 0.2105 $\pm$ 0.1372	& 0.09 & 	0.0721 $\pm$	0.0983 & 0.06\\ \cline{2-6}
 & 1000 & 0.0766 $\pm$ 0.0608 &	0.86 & 0.0240	$\pm$ 0.0182	& 0.53
 \\
 \cline{2-6}
 & 10000 & 0.0282 $\pm$	0.0181 & 8.64 & 0.0091 $\pm$ 0.0069	& 5.53 \\ \cline{2-6}
 & 100000 & 0.0086 $\pm$ 0.0067 & 89.93	& 0.0034 $\pm$ 0.0027 & 57.64\\
\midrule
\midrule
\multirow{4}{*}{\textit{Andes}} & 100 & 0.0821 $\pm$ 0.0477 & 1.07 &	0.0619 $\pm$ 0.0453 & 0.22 \\ \cline{2-6}
 & 1000 & 0.0257 $\pm$ 0.0184 & 10.62 & 0.0163 $\pm$ 0.0139 & 2.20 \\ \cline{2-6}
 & 10000 & 0.0087 $\pm$	0.0069 & 106.55 & 0.0058 $\pm$ 0.0042 &	22.62\\ \cline{2-6} 
 & 100000 & 0.0025 $\pm$ 0.0015 & 1074.93 & 0.0020 $\pm$ 0.0016 & 233.72\\
\bottomrule
\end{tabular}
\end{adjustbox}
\caption{The mean absolute error (MAE), the standard deviation of the error (Std.), and the average elapsed time (in seconds) versus the number of samples (N). For each case, LW and CS-LW were executed 30 times.}
\label{Table: Result1}
\end{table}

We used the standard decision tree learning algorithm to detect structures and overfitted it on tabular-CPDs to get tree-CPDs, which was then converted into rules.
Let us denote the program with these rules by $\mathbb{P}_{tree}$. CS-LW is implemented in the Prolog programming language, thus to compare the sampling speed of LW with CS-LW, we need a similar implementation of LW. Fortunately, we can use the same implementation of CS-LW for obtaining LW estimates. Recall that if we do not make structures explicit in rules and represent each entry in tabular-CPDs with rules, then CS-LW boils down to LW. Let $\mathbb{P}_{table}$ denotes the program where each rule in it corresponds to an entry in tabular-CPDs. Table \ref{Table: Result1} shows the comparison of estimates obtained using $\mathbb{P}_{tree}$ (CS-LW) and $\mathbb{P}_{table}$ (LW). Note that CS-LW automatically discards non-requisite variables for sampling. So, we chose the query and evidence such that almost all variables in BNs were requisite for the conditional query. 

\begin{table*}[t]
\centering
\begin{adjustbox}{width=2.0\columnwidth,center}
\begin{tabular}{|c|cc|cc|cc|cc|cc|}
\toprule
         &   \multicolumn{2}{|c|}{\textbf{LW}} &   \multicolumn{2}{|c|}{\textbf{CC-10,000}} & \multicolumn{2}{|c|}{\textbf{CC-100,000}} & \multicolumn{2}{|c|}{\textbf{CC-1000,000}} & \multicolumn{2}{|c|}{\textbf{CS-LW}}\\ \hline\hline
\textbf{BN} & \textbf{N} & \textbf{MAE $\pm$ Std.} & \textbf{N} & \textbf{MAE $\pm$ Std.} & \textbf{N} & \textbf{MAE $\pm$ Std.} & \textbf{N} & \textbf{MAE $\pm$ Std.} & \textbf{N} & \textbf{MAE $\pm$ Std.}\\
\midrule
\textit{Alarm} & 131606 & 0.0073 $\pm$ 0.0054 & 3265 &	0.0022 $\pm$ 0.0018 & NA & 0 $\pm$ 0 (exact) & NA & 0 $\pm$ 0 (exact) & 178620 & 0.0019 $\pm$ 0.0016 \\ \hline
\textit{Win95pts} & 51956 & 0.0022 $\pm$ 0.0016 & 635 &	0.0149 $\pm$ 0.0163 & NA & 0 $\pm$ 0 (exact) & NA & 0 $\pm$ 0 (exact) & 67855 & 0.0017 $\pm$ 0.0011\\ \hline
\textit{Andes} & 13113 & 0.0068 $\pm$ 0.0062 &116 & 0.0814 $\pm$ 0.0915 & 17 & 0.0060 $\pm$ 0.0094 & NA & 0 $\pm$ 0 (exact) & 56672 & 	0.0022 $\pm$ 0.0021\\ \hline
\textit{Munin1} & 15814 & 0.0036 $\pm$ 0.0026 & \multicolumn{2}{|c|}{out of memory}	& \multicolumn{2}{|c|}{out of memory} & \multicolumn{2}{|c|}{out of memory} & 17985 &	0.0035 $\pm$ 0.0025\\
\bottomrule
\end{tabular}
\end{adjustbox}
\caption{The mean absolute error (MAE), the standard deviation of the error (Std.), and the average number of samples (N) drawn when algorithms were run 50 times for 2 minutes (approx.) each.
The algorithms are: LW, CS-LW, CC with circuit size 10,000, with size 100,000, and with size 1000,000.}
\label{Table: Result2}
\end{table*}

As expected, we observe that less time is required by CS-LW to generate the same number of samples. This is because it visits only the subset of requisite variables in each simulation. \textit{Andes} has more structures compared to \textit{Alarm}. Thus, the sampling speed of CS-LW is much faster compared to LW in \textit{Andes}. Additionally, we observe that the estimate, with the same number of samples, obtained by CS-LW is much better than LW. This is significant. It is worth emphasizing that approaches based on collapsed sampling obtain better estimates than LW with the same number of samples, but then the speed of drawing samples significantly decreases. 
In CS-LW, the speed increases when structures are present. This is possible because CS-LW exploits CSIs. 

Hence, we get the answer to the first two questions: When many structures are present, and when they are made explicit in rules, then CS-LW will draw samples faster compared to LW. Additionally, estimates will be better with the same number of samples.

To answer our final question, we compared CS-LW with the collapsed compilation \citep[CC,][]{friedman2018approximate}, which has been recently shown to outperform several sampling algorithms.
It combines a state of the art exact inference algorithm that exploits CSIs and importance sampling that scales the exact inference. 
The load between the exact and sampling can be regulated using the size of the arithmetic circuit: larger the circuit's size, larger the load on the exact and lesser the load on the sampling, i.e., less variables are considered for sampling. For this experiment, we consider two additional BNs: i) \textit{Win95pts}, a system for printing troubleshooting in Windows 95 with 76 variables; ii) \textit{Munin1}, an expert EMG assistant with 186 variables.
However, not many structures are present in the CPDs of these two BNs, so not much difference in the performance of LW and CS-LW is expected. 

The comparison is shown in Table \ref{Table: Result2}. We can observe the following: i) as expected from collapsed sampling, much fewer samples are drawn in the same time; ii) the right choice of circuit's size is crucial, e.g., with circuit size 10,000, CC performs poorly compared to LW on some BNs while better when the size is increased; iii) CS-LW performs better compared to CC when the circuit is not huge; iv) on the three BNs, CC with a huge circuit size computes the exact conditional probability while LW and CS-LW can only provide a good approximation of that in the same time.

To demonstrate that the fourth observation does not undermine the importance of pure sampling, we used \textit{Munin1}. Although the size of this BN is comparable to the size of \textit{Andes}, almost all variables are multi-valued, and their domain size can be as large as 20; hence, some CPDs are huge, while in \textit{Andes}, variables are binary-valued. CC that works well on \textit{Andes}, fails to deal with huge CPDs of \textit{Munin1} on a machine with $16$ GB memory. On the other hand, both LW and CS-LW work well on this BN. 

Hence, we get the answer to our final question: CS-LW is competitive with the state-of-the-art and can be a useful tool for inference on massive BNs with structured CPDs.

\section{Related Work}
Although the question of how to exploit CSIs arising due to structures 
is not new and has puzzled researchers for decades, research in this direction has mainly been focused on exact inference \citep{boutilier1996context, zhang1999role, poole1997probabilistic, poole2003exploiting}. Nowadays, it is common to use knowledge compilation (KC) based exact inference for the purpose \citep{chavira2008probabilistic, fierens2015inference, shen2016tractable}. There are not many approximate inference algorithms that can exploit them. However, there are some tricks that make use of structures to approximate the probability of a query. One trick, introduced by \cite{10.5555/2074094.2074147}, is to make rule-base simpler by ignoring distinctions in close probabilities. Another trick, explored in the case of the tree-CPDs, is to prune trees and reduce the size of actual CPDs \citep{salmeron2000importance,cano2011approximate}. However, approximation by making distribution simpler is orthogonal to traditional ways of approximation, such as sampling. \cite{fierens2010context} observed the speedup in Gibbs sampling due to structures; however, did not consider global implications of structures. 

Recently, \cite{friedman2018approximate} realized that KC is good at exploiting structures while sampling is scalable; thus, proposed CC that inherits advantages of both. However, along with advantages, this approach also inherits the scalability limitations of KC. Furthermore, CC is limited to discrete distributions. The problem of exploiting CSIs in discrete-continuous distributions is non-trivial and is poorly studied. Recently, it has attracted some attention \citep{zeng2019efficient}. However, proposed approaches are also exact and rely on complicated weighted model integration \citep{belle2015probabilistic}, which quickly become infeasible. CS-LW is simple, scalable, and applies to such distributions. A sampling algorithm for a rule-based representation of discrete-continuous distributions was developed by \cite{nitti2016probabilistic}; however, it did not exploit CIs and global implications of rule structures.

\section{Conclusion}
We studied the role of CSI in approximate inference and introduced a notion of contextual assignments to show that CSIs allow for breaking the main problem of estimating conditional probability query into several small problems that can be estimated independently. Based on this notion, we presented an extension of LW, which not only generates samples faster; it also provides a better estimate of the query with much fewer samples. Hence, we provided a solid reason to use structured-CPDs over tabular-CPDs. Like LW, we believe other sampling algorithms can also be extended along the same line. We aim to open up a new direction towards improved sampling algorithms that also exploit CSIs.

\paragraph{Acknowledgements} This work has received funding from the European Research Council (ERC) under the European Union’s Horizon 2020 research and innovation programme (grant agreement No [694980] SYNTH: Synthesising Inductive Data Models). OK was supported by the Czech Science Foundation project ``Generative Relational Models'' (20-19104Y) and partially also by the OP VVV project {\it CZ.02.1.01/0.0/0.0/16\_019/0000765} ``Research Center for Informatics''. The authors would like to thank Luc De Raedt, Jessa Bekker, Pedro Zuidberg Dos Martires and the anonymous reviewers for valuable feedback.


\bibliographystyle{unsrtnat}
\bibliography{biblio}

\onecolumn
\aistatstitle{Context-Specific Likelihood Weighting: \\
Supplementary Materials}

\aistatsauthor{ Nitesh Kumar \And Ond\v rej Ku\v zelka}
\aistatsaddress{ Department of Computer Science and Leuven.AI \\KU Leuven, Belgium \And  Department of Computer Science \\ Czech Technical University in Prague, Czechia}

\section{Missing Proofs}
\subsection{Proof of Lemma \ref{theorem: bayes-ball 1}}
In this section, we present the detailed proof of Lemma  \ref{theorem: bayes-ball 1}. 

\begin{proof}
Let us denote the variables in $\mathbf{Z}$ that are marked on the top (requisite) by $\mathbf{Z}_\star$ and that are not marked on the top (not requisite) by $\mathbf{Z}_{\bar{\star}}$. The required probability $\mu$ is then given by,
\begin{equation*}
    \begin{aligned}
    \mu = P(\mathbf{x}_q \mid \mathbf{e}) = 
    \frac{\sum_{\mathbf{x}, \mathbf{z}_{\star}, \mathbf{z}_{\bar{\star}}} P(\mathbf{x},  \mathbf{z}_{\star}, \mathbf{z}_{\bar{\star}}, \mathbf{e}) f(\mathbf{x})}{\sum_{\mathbf{x}, \mathbf{z}_{\star}, \mathbf{z}_{\bar{\star}}} P(\mathbf{x},  \mathbf{z}_{\star}, \mathbf{z}_{\bar{\star}}, \mathbf{e})} = \frac{\sum_{\mathbf{x}, \mathbf{z}_\star} P(\mathbf{x},  \mathbf{z}_\star, \mathbf{e}) f(\mathbf{x}) \sum_{\mathbf{z}_{\bar{\star}}} P(\mathbf{z}_{\bar{\star}} \mid \mathbf{x},  \mathbf{z}_\star, \mathbf{e}) }{\sum_{\mathbf{x}, \mathbf{z}_\star} P(\mathbf{x},  \mathbf{z}_\star, \mathbf{e}) \sum_{\mathbf{z}_{\bar{\star}}} P(\mathbf{z}_{\bar{\star}} \mid \mathbf{x},  \mathbf{z}_\star, \mathbf{e})}
    \end{aligned}
\end{equation*}
Since $\sum_{\mathbf{z}_{\bar{\star}}} P(\mathbf{z}_{\bar{\star}} \mid \mathbf{x},  \mathbf{z}_\star, \mathbf{e}) = 1$, we can write,
\begin{equation*}
    \begin{aligned}
    \mu = \frac{\sum_{\mathbf{x}, \mathbf{z}_\star} P(\mathbf{x},  \mathbf{z}_\star, \mathbf{e}) f(\mathbf{x})}{\sum_{\mathbf{x}, \mathbf{z}_\star} P(\mathbf{x},  \mathbf{z}_\star, \mathbf{e})}
    \end{aligned}
\end{equation*}
Now let us denote the observed variables in $\mathbf{E}$ that are visited (requisite) by $\mathbf{E}_r$ and those that are not visited (not requisite) by $\mathbf{E}_n$. We can write, 
\begin{equation*}
    \begin{aligned}
    \mu = \frac{\sum_{\mathbf{x}, \mathbf{z}_\star} P(\mathbf{x},  \mathbf{z}_\star, \mathbf{e}_r) P(\mathbf{e}_n \mid \mathbf{x},  \mathbf{z}_\star, \mathbf{e}_r) f(\mathbf{x})}{\sum_{\mathbf{x}, \mathbf{z}_\star} P(\mathbf{x},  \mathbf{z}_\star, \mathbf{e}_r) P(\mathbf{e}_n \mid \mathbf{x},  \mathbf{z}_\star, \mathbf{e}_r)}
    \end{aligned}
\end{equation*}
The variables in $\mathbf{X} \cup \mathbf{Z}_\star$ pass the Bayes-balls to all their parents and all their children, but $\mathbf{E}_n$ is not visited by these balls. The correctness of the Bayes-ball algorithm ensures that there is no active path from $\mathbf{X} \cup \mathbf{Z}_\star$ to any $E_n$ in $\mathbf{E}_n$ given $\mathbf{E}_r$. Thus $\mathbf{X}, \mathbf{Z}_\star \perp \mathbf{E}_n \mid \mathbf{E}_r$ and $P(\mathbf{e}_n \mid \mathbf{x},  \mathbf{z}_\star, \mathbf{e}_r) = P(\mathbf{e}_n \mid \mathbf{e}_r)$. After cancelling out the common term $P(\mathbf{e}_n \mid \mathbf{e}_r)$, we get,
\begin{equation*}
    \begin{aligned}
    \mu = \frac{\sum_{\mathbf{x}, \mathbf{z}_\star} P(\mathbf{x},  \mathbf{z}_\star, \mathbf{e}_r) f(\mathbf{x})}{\sum_{\mathbf{x}, \mathbf{z}_\star} P(\mathbf{x},  \mathbf{z}_\star, \mathbf{e}_r)}
    \end{aligned}
\end{equation*}
Now let us denote observed variables in $\mathbf{E}_r$ that are only visited by $\mathbf{E}_\smwhitestar$ and that are visited as well as marked on top by $\mathbf{E}_\star$. After cancelling out the common term, we get the desired result, 
\begin{equation*}
    \begin{aligned}
    \mu = \frac{\sum_{\mathbf{x}, \mathbf{z}_\star} P(\mathbf{x},  \mathbf{z}_\star, \mathbf{e}_\star \mid \mathbf{e}_\smwhitestar) P(\mathbf{e}_\smwhitestar) f(\mathbf{x})}{\sum_{\mathbf{x}, \mathbf{z}_\star} P(\mathbf{x},  \mathbf{z}_\star, \mathbf{e}_\star \mid \mathbf{e}_\smwhitestar) P(\mathbf{e}_\smwhitestar)}  = 
    \frac{\sum_{\mathbf{x}, \mathbf{z}_\star} P(\mathbf{x},  \mathbf{z}_\star, \mathbf{e}_\star \mid \mathbf{e}_\smwhitestar) f(\mathbf{x})}{\sum_{\mathbf{x}, \mathbf{z}_\star} P(\mathbf{x},  \mathbf{z}_\star, \mathbf{e}_\star \mid \mathbf{e}_\smwhitestar)}
    \end{aligned}
\end{equation*}
\end{proof}

\begin{example}
Consider the network of Figure \ref{fig:context-specific independence}, and assume that our evidence is $\{D=1, F=1, G=0, H=1\}$, and our query is $\{E=0\}$. Suppose we start by visiting the query variable from its child and apply the four rules of Bayes-ball. One can easily verify that observed variables $F, G, H$ will be marked on top; hence $\{F=1, G=0, H=1\}$ is diagnostic evidence ($\mathbf{e}_\star$). The observed variable $D$ will only be visited; hence $\{D=1\}$ is predictive evidence ($\mathbf{e}_\smwhitestar$). Variables $A,B,C,E$ will be marked on top and are requisite unobserved variables ($\mathbf{X} \cup \mathbf{Z}_\star$).  
\end{example}

\subsection{Proof of Theorem \ref{Theorem: the expecation}}
In this section, we present the detailed proof of Theorem \ref{Theorem: the expecation}. 

\begin{proof}
The expectation $\mathbb{E}_{Q_\star}[W_{\mathbf{\dot{e}}_\star}]$ is given by
\begin{equation*}
\begin{aligned}
    \sum_{\mathbf{x}, \mathbf{z}_\star} \prod_{u_i \in \mathbf{x} \cup \mathbf{z}_\star} P(u_i \mid \mathbf{pa}(U_i)) \prod_{v_i \in \mathbf{\dot{e}}_\star} P(v_i \mid \mathbf{pa}(V_i)).
\end{aligned}
\end{equation*}
The basis $\mathbf{\dot{S}}_\star$ is a subset of $\mathbf{X} \cup \mathbf{Z}_\star$ by Definition \ref{definition: Basis} . Let us denote $(\mathbf{X} \cup \mathbf{Z}_\star) \setminus \mathbf{\dot{S}}_\star$ by $\mathbf{Z}_{\diamond}$. We can now rewrite the expectation as follows, 
\begin{equation*}
\begin{aligned}
    \sum_{\mathbf{\dot{s}}_\star, \mathbf{z}_{\diamond}} \prod_{u_i \in \mathbf{\dot{e}}_\star \cup \mathbf{\dot{s}}_\star} P(u_i \mid \mathbf{pa}(U_i)) \prod_{v_i \in \mathbf{z}_\diamond} P(v_i \mid \mathbf{pa}(V_i)).
\end{aligned}
\end{equation*}
We will show that $Pa \notin \mathbf{Z}_{\diamond}$ for any $Pa \in \mathbf{Pa}(U_i)$, which will then allow us to push the summation over $\mathbf{z}_{\diamond}$ inside. Let us consider two cases: 
\begin{itemize}
    \item For $U_i \in \mathbf{\dot{E}}_\star$, let $Pa \in \mathbf{Pa}(U_i)$ be an unobserved parent of $U_i$, then there will be a direct causal trail from $Pa$ to $U_i$, consequently $Pa$ will be in the set $\mathbf{\dot{S}}_{\star}$.
    \item For $U_i \in \mathbf{\dot{S}}_{\star}$, there will be a causal trail $U_i \rightarrow \cdots\ B_j\ \cdots \rightarrow E$ such that $E \in \mathbf{\dot{E}}_\star$ and such that either no $B_i$ is observed or there is no $B_i$. Let $Pa \in \mathbf{Pa}(U_i)$ be an unobserved parent of $U_i$ then there will be a direct causal trail from $Pa$ to $U_i$, consequently, there will be such causal trail from $Pa$ to $E$ and $Pa$ will be in the set $\mathbf{\dot{S}}_{\star}$.
\end{itemize}
Hence, we push the summation over  $\mathbf{z}_{\diamond}$ inside and use the fact that $\sum_{\mathbf{z}_{\diamond}} \prod_{v_i \in \mathbf{z}_{\diamond}} P(v_i \mid \mathbf{pa}(V_i)) = 1$, to get the desired result.  
\end{proof}

\subsection{Proof of Theorem \ref{theorem: cslw_main}}
In this section, we present the detailed proof of Theorem \ref{theorem: cslw_main}. 
\begin{proof}
Since $\mathbf{X}, \mathbf{Z}_\star, \mathbf{E}_\star, \mathbf{E}_\smwhitestar$ are variables of the Bayesian network $\mathcal{B}$ and they form a sub-network $\mathcal{B}_\star$ such that $\mathbf{E}_\smwhitestar$ do not have any parent, we can always write,
\begin{equation*}\label{equation: theorem4-1}
\begin{aligned}
    P(\mathbf{x}, \mathbf{z}_\star, \mathbf{e}_\star \mid \mathbf{e}_\smwhitestar) = \prod_{u_i \in \mathbf{x} \cup \mathbf{z}_\dagger \cup \mathbf{e}_\dagger} P(u_i \mid \mathbf{pa}(U_i)) \prod_{v_i \in \mathbf{z}_\ddagger \cup \mathbf{e}_\ddagger} P(v_i \mid \mathbf{pa}(V_i))
\end{aligned}
\end{equation*}
such that $p \in \mathbf{x} \cup \mathbf{z}_\star \cup \mathbf{e}_\star \cup \mathbf{e}_\smwhitestar$ for all $p \in \mathbf{pa}(U_i)$ or $p \in \mathbf{pa}(V_i)$. Now consider the summation over all possible assignments of variables in $\mathbf{X}, \mathbf{Z}_\star$, that is: $\sum_{\mathbf{x}, \mathbf{z}_\star} P(\mathbf{x}, \mathbf{z}_\star, \mathbf{e}_\star \mid \mathbf{e}_\smwhitestar)$. 
We can always write, 
\begin{equation}\label{equation: theorem4-2}
\begin{aligned}
    \sum_{\mathbf{x}, \mathbf{z}_\star} P(\mathbf{x}, \mathbf{z}_\star, \mathbf{e}_\star \mid \mathbf{e}_\smwhitestar) = \sum_{\psi \in \Psi}\sum_{\mathbf{z}_\ddagger[\psi]} P(\mathbf{x}[\psi], \mathbf{z}_\dagger[\psi], \mathbf{z}_\ddagger[\psi],  \mathbf{e}_\dagger[\psi], \mathbf{e}_\ddagger[\psi] \mid \mathbf{e}_\smwhitestar)
\end{aligned}
\end{equation}
To simplify notation, from now we denote $\{ \mathbf{x}[\psi]$, $\mathbf{z}_\dagger[\psi]$, $\mathbf{Z}_\ddagger[\psi]$,  $\mathbf{e}_\dagger[\psi]$, $\mathbf{e}_\ddagger[\psi] \}$ by $\{ \mathbf{x}$, $\mathbf{z}_\dagger$, $\mathbf{Z}_\ddagger$,  $\mathbf{e}_\dagger$, $\mathbf{e}_\ddagger \}$. After using the definition of contextual assignments, we have that, 
\begin{equation*}
\begin{aligned}
    P(\mathbf{x}, \mathbf{z}_\dagger, \mathbf{z}_\ddagger,  \mathbf{e}_\dagger, \mathbf{e}_\ddagger \mid \mathbf{e}_\smwhitestar)  = \prod_{u_i \in \mathbf{x} \cup \mathbf{z}_\dagger \cup \mathbf{e}_\dagger} P(u_i \mid \mathbf{ppa}(U_i)) \prod_{v_i \in \mathbf{z}_\ddagger \cup \mathbf{e}_\ddagger} P(v_i \mid \mathbf{pa}(V_i))
\end{aligned}
\end{equation*}
Since $p \notin \mathbf{z}_\ddagger$ for any $p \in \mathbf{ppa}(U_i)$, we can push the summation over $\mathbf{z}_\ddagger$ inside to get, 
\begin{equation}\label{equation: theorem4-3}
\begin{aligned}
    \sum_{\psi \in \Psi}\sum_{\mathbf{z}_\ddagger} P(\mathbf{x}, \mathbf{z}_\dagger, \mathbf{z}_\ddagger,  \mathbf{e}_\dagger, \mathbf{e}_\ddagger \mid \mathbf{e}_\smwhitestar) = \sum_{\psi \in \Psi}\prod_{u_i \in \mathbf{x} \cup \mathbf{z}_\dagger \cup \mathbf{e}_\dagger} P(u_i \mid \mathbf{ppa}(U_i)) \sum_{\mathbf{z}_\ddagger}\prod_{v_i \in \mathbf{z}_\ddagger \cup \mathbf{e}_\ddagger} P(v_i \mid \mathbf{pa}(V_i)).
\end{aligned}
\end{equation}
However, we get a strange term $\sum_{\mathbf{z}_\ddagger}\prod_{v_i \in \mathbf{z}_\ddagger \cup \mathbf{e}_\ddagger} P(v_i \mid \mathbf{pa}(V_i))$. 
Let $\mathbf{S}_{\ddagger}$ denote the basis of residual $\mathbf{e}_\ddagger$. We have that $\mathbf{S}_{\ddagger} \subseteq \mathbf{Z}_\ddagger$ by Definition \ref{Definition: safe contexual assignments}. Let us denote $\mathbf{Z}_\ddagger \setminus \mathbf{S}_{\ddagger}$ with $\mathbf{Z}_{\diamond}$. Now the strange term can be rewritten as,
\begin{equation*}
\begin{aligned}
    \sum_{\mathbf{s}_{\ddagger}, \mathbf{z}_{\diamond}} \prod_{u_i \in \mathbf{e}_\ddagger \cup \mathbf{s}_{\ddagger}} P(u_i \mid \mathbf{pa}(U_i)) \prod_{v_i \in \mathbf{z}_{\diamond}} P(v_i \mid \mathbf{pa}(V_i)).
\end{aligned}
\end{equation*}
In the proof of Theorem \ref{Theorem: the expecation}, we showed that the summation over variables not in $\mathbf{S}_\ddagger$ can be pushed inside; hence, $\mathbf{Z}_{\diamond}$ can be pushed inside. After using the fact that $\sum_{\mathbf{z}_{\diamond}} \prod_{v_i \in \mathbf{z}_{\diamond}} P(v_i \mid \mathbf{pa}(V_i)) = 1$, we conclude that the strange term is actually the expectation $\mathbb{E}_{Q_\star}[W_{\mathbf{e}_\ddagger}]$. Using Equation (\ref{equation: bb}), (\ref{equation: theorem4-2}), (\ref{equation: theorem4-3}) and rearranging terms, the result follows. 
\end{proof}

\subsection{Proof of Lemma \ref{theorem: partial assignments}}
In this section, we present the detailed proof of Lemma  \ref{theorem: partial assignments}. 
\begin{proof}
It is clear that a subset of unobserved variables is assigned. Let $\mathbf{Z}_\ddagger$ be a set of unobserved variables left unassigned. Let $E \in \mathbf{E}_\star$ be an observed variable. Consider two cases: 
\begin{itemize}
    \item All ancestors of $E$ are in  $\mathbf{Z}_\ddagger \cup \mathbf{E}_\smwhitestar \cup \mathbf{E}_\star$.
    \item Some ancestors of $E$ are in $\mathbf{Z}_\ddagger \cup \mathbf{E}_\smwhitestar \cup \mathbf{E}_\star$ and some are in $\mathbf{X} \cup \mathbf{Z}_\dagger$. Let $A \in \mathbf{X} \cup \mathbf{Z}_\dagger$ and let $A \rightarrow \cdots\ B_i\ \cdots \rightarrow E$ be a causal trail.  Some $B_i$ are observed in all such trails. 
\end{itemize}
Clearly, $E$ will not be visited from any parent in the first case, and in the second case, the visit will be blocked by observed variables. Consequently, $E$ will not be weighted, which completes the proof. 
\end{proof}

\subsection{Proof of Theorem \ref{theorem: dc-partial-justification}}
In this section, we present the detailed proof of Theorem \ref{theorem: dc-partial-justification}. 
\begin{proof}
Variables in $\mathbf{Z}_\ddagger$ are not assigned in the simulation; hence, it follows immediately from Lemma \ref{theorem: DC CSI} that the assignment is contextual. Assume by contradiction that $A \in \mathbf{X} \cup \mathbf{Z}_\dagger$, $E \in \mathbf{E}_\ddagger$ and there is a causal trail $A \rightarrow \cdots\ B_i\ \cdots \rightarrow E$ such that no $B_i$ is observed or there is no $B_i$. Since $ A $ is assigned, all children of $ A $ will be visited, and following the trail, the variable $ E $ will also be visited from its parent since there is no observed variable in the trail to block the visit. Consequently, $E$ will be weighted, which contradicts our assumption that $E$ is not weighted. Hence, the assignment is also safe.
\end{proof}

\subsection{Proof of Lemma \ref{theorem: DC CSI}}
In this section, we present the detailed proof of Lemma \ref{theorem: DC CSI}.
\begin{proof}
Since $A$ is assigned/weighted and rules in $\mathbb{P}$ are exhaustive, a rule $\mathcal{R} \in \mathbb{P}$ with $A$ in its head must have fired. Let $\mathbf{d}$ be a body and $\mathcal{D}$ be a distribution in the head of $\mathcal{R}$. Since each $d_i \in \mathbf{d}$ must be true for $\mathcal{R}$ to fire, $\mathbf{d} \subseteq \mathbf{c}$. We assume that rules in $\mathbb{P}$ are mutually exclusive. Thus, among all rules for $A$, only $\mathcal{R}$ will fire even when an assignment of some variables in $\mathbf{B}$ is also given. Hence, by definition of the rule $\mathcal{R}$,
we have that,
\begin{equation*}
    \mathcal{D} = P(A \mid \mathbf{d}) = P(A \mid \mathbf{c}) = P(A \mid \mathbf{c}, \mathbf{B})
\end{equation*}
\end{proof}

\vskip -0.11in
\section{Top-Down Proof Procedure for DC($\mathcal{B}$)}\label{Section: DC Proofs}
\begin{figure}[t]
    \centering
    \includegraphics[width=1\linewidth]{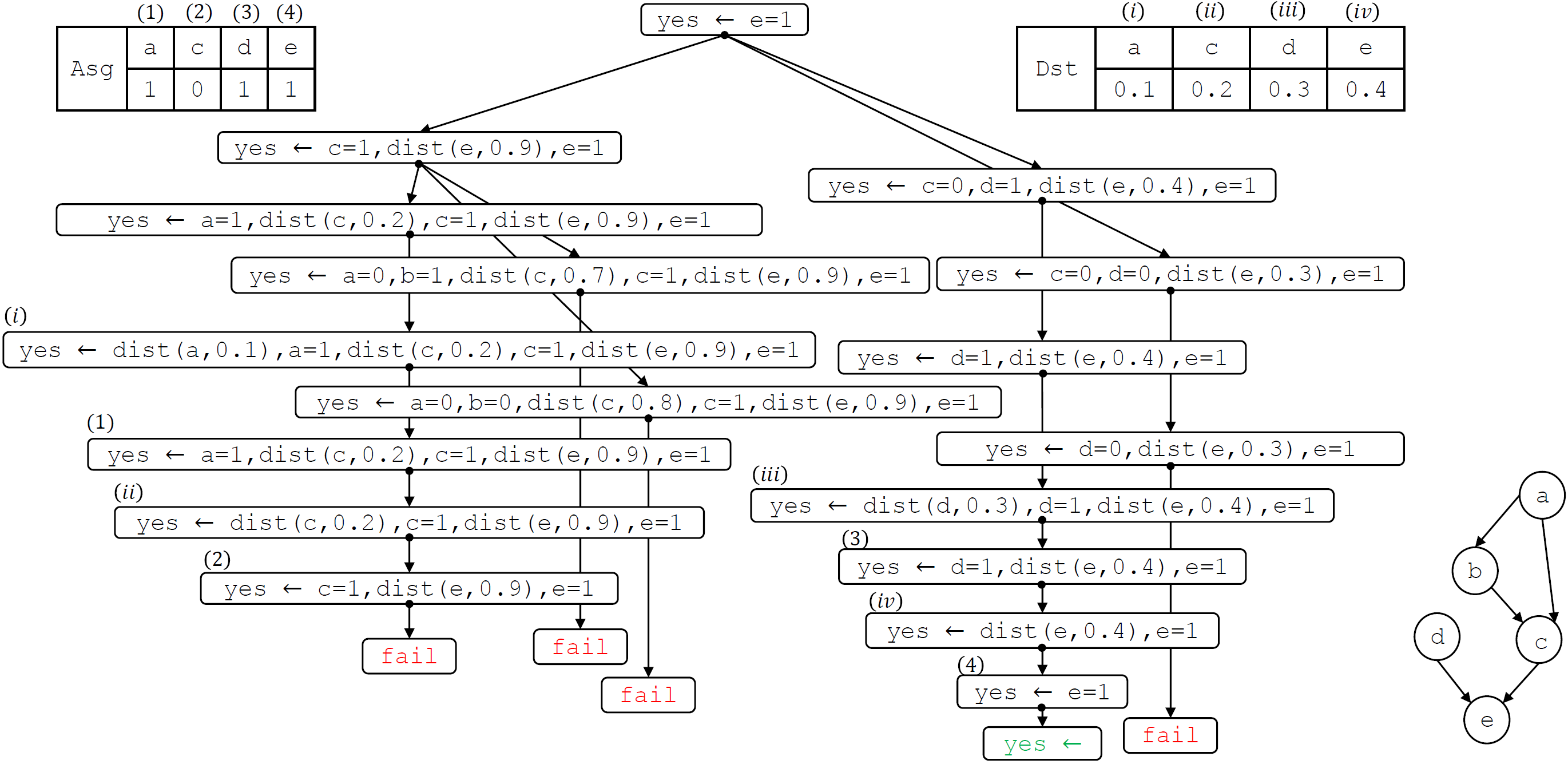}
    \caption{Left: A search graph induced to prove \texttt{e=1}; Right: A graphical structure.}
    \label{fig: proof}
\end{figure}

Let us look into the process of estimating the unconditional probability of queries to DC($\mathcal{B}$) programs. We assume some familiarity with proof procedures for definite clauses \citep{poole2010artificial}.


Just like the set of definite clauses forms the knowledge base, the DC($\mathcal{B}$) program forms a {\em probabilistic knowledge base}. We can ask queries of the form \texttt{yes $\leftarrow$ e=1}, which is a question: is \texttt{e} assigned to \texttt{1}? We first need to prove that \texttt{e=1} before concluding that the answer is \texttt{yes}. 
To realize that we perform a proof procedure, from the query, to determine whether it is a logical consequence of rules in the DC($\mathcal{B}$) program. Algorithm \ref{algorithm: proof} describes the procedure, which is similar to the standard SLD-resolution for definite clauses. However, there are some differences to prove atoms of the form \texttt{e=1} due to the stochastic nature of sampling. We illustrate the proof procedure with an example.
\begin{example}\label{example: proof process}
\normalfont Consider a Bayesian network whose graph structure is as shown in Figure \ref{fig: proof} (right) and whose CPDs are expressed using the following rules:
\begin{lstlisting}[frame=none]
a ~ bernoulli(0.1).
d ~ bernoulli(0.3).
b ~ bernoulli(0.2) := a=0.
b ~ bernoulli(0.6) := a=1.
c ~ bernoulli(0.2) := a=1.
c ~ bernoulli(0.7) := a=0,,b=1.
c ~ bernoulli(0.8) := a=0,,b=0.
e ~ bernoulli(0.9) := c=1.
e ~ bernoulli(0.4) := c=0,,d=1.
e ~ bernoulli(0.3) := c=0,,d=0.\end{lstlisting} Suppose the query \texttt{yes $\leftarrow$ e=1} is asked. The procedure induces a search graph. An example of such a graph is shown in Figure \ref{fig: proof} (left),  where we write \texttt{dist(y, p)} for \texttt{dist(y, bernoulli(p))} and use comma(\texttt{,}) instead of \texttt{$\wedge$} since there is no risk of confusion. In this example, the proof succeeds using a derivation. However, it might happen that the proof can not be derived. In that case, the proof fails, and the answer is \texttt{no}. 
\end{example}
After repeating the procedure, the fraction of times we get \texttt{yes} is the estimated probability of the query. It is important to note that some requisite variables may not be assigned in some occasions, e.g., in the proof shown in Figure \ref{fig: proof}, variable \texttt{b}, which is requisite to compute the probability, is not assigned. Hence, we sample values of \texttt{e} faster. In this way, the procedure exploits the structure of rules.  
\section{Additional Experimental Details}\label{appendix: additional exp details}

To make sure that almost all variables in BNs are requisite, we used the following query variables and evidence to obtain  the results (Table \ref{Table: Result1} and Table \ref{Table: Result2}): 
\begin{itemize}
    \item \textit{Alarm}
    \begin{itemize}
        \item $\mathbf{x} = $ \textit{\{bp = low\}}
        \item $\mathbf{e} = $ \textit{\{lvfailure = false, cvp = normal, hr = normal, expco2 = low, ventalv = low, ventlung = zero\}}
    \end{itemize}
    \item \textit{Win95pts}
    \begin{itemize}
        \item $\mathbf{x} = $ \textit{\{problem1 = normal\_output\}}
        \item $\mathbf{e} = $ \textit{\{prtstatoff = no\_error, prtfile = yes, prtstattoner = no\_error, repeat = yes\_\_always\_the\_same\_, ds\_lclok = yes, lclok = yes, problem3 = yes, problem4 = yes, nnpsgrphc = yes, psgraphic = yes, problem5 = yes, gdiin = yes, appdata = correct, prtstatpaper = no\_error\}}
    \end{itemize}
    \item \textit{Andes}
    \begin{itemize}
        \item $\mathbf{x} = $ \textit{\{grav78 = false\}}
        \item $\mathbf{e} = $ \textit{\{goal\_2 = true, displacem0 = false, snode\_10 = true, snode\_16 = true, grav2 = true, constant5 = false, known8 = true, try11 = true, kinemati17 = false, try13 = true, given21 = true, choose35 = false, write31 = false, need36 = false, resolve38 = true, goal\_69 = false, snode\_73 = false, goal\_79 = false, try24 = false, newtons45 = false, try26 = false, snode\_65 = false, snode\_88 = false, buggy54 = true, weight57 = true, goal\_104 = false, goal\_108 = false, need67 = false, goal\_114 = false, snode\_118 = false, snode\_122 = false, snode\_125 = false, goal\_129 = false, snode\_135 = false, goal\_146 = false, snode\_151 = false\}}
    \end{itemize}
    \item \textit{Munin1}
    \begin{itemize}
        \item $\mathbf{x} = $ \textit{\{rmedd2amprew = r04\}}
        \item $\mathbf{e} = $ \textit{\{rlnlt1apbdenerv = no, rlnllpapbmaloss = no, rlnlwapbderegen = no, rdiffnapbmaloss = no, rderegenapbnmt = no, rdiffnmedd2block = no, rmeddcvew = ms60, rapbnmt = no, rapbforce = 5, rlnlbeapbdenerv = no, rlnlbeapbneuract = no, rmedldwa = no, rmedd2blockwd = no\}}
    \end{itemize}
\end{itemize}

\section{Code}
The source code is present in the supplementary material along with installation instructions.

\end{document}